%% file: Formatting-Instructions-LaTeX-2026.tex
\documentclass[letterpaper]{article} 
\usepackage{aaai2026}  
\usepackage{times}  
\usepackage{helvet}  
\usepackage{courier}  
\usepackage[hyphens]{url}  
\usepackage{graphicx} 
\usepackage{natbib}  
\usepackage{caption} 
\usepackage{algorithm}
\usepackage{algorithmic}
\usepackage{booktabs}
\usepackage{xcolor}
\usepackage{amsmath}
\usepackage{pifont}
\usepackage{amssymb}
\definecolor{darkgreen}{rgb}{0.15, 0.75, 0.15}
\definecolor{cvprblue}{rgb}{0.21,0.49,0.74}
\definecolor{lightblue}{rgb}{0.90, 0.95, 0.99}
\definecolor{pinkishred}{rgb}{0.9, 0.4, 0.4}
\definecolor{lightblue}{rgb}{0.678, 0.847, 0.902}
\usepackage{newfloat}
\usepackage{listings}
\DeclareCaptionStyle{ruled}{labelfont=normalfont,labelsep=colon,strut=off} 
\floatstyle{ruled}
\newfloat{listing}{tb}{lst}{}
\floatname{listing}{Listing}
\title{MCIE: Multimodal LLM-Driven Complex Instruction \\ Image Editing with Spatial Guidance}
\author{
    Xuehai Bai\textsuperscript{\rm 1}, Xiaoling Gu\textsuperscript{\rm 1}\thanks{Corresponding author.}, Akide Liu\textsuperscript{\rm 2}, Hangjie Yuan\textsuperscript{\rm 3},YiFan Zhang\textsuperscript{\rm 4}, Jack Ma\textsuperscript{\rm 5}\\
}
\affiliations{
    \textsuperscript{\rm 1}Hangzhou Dianzi University,  \\
    \textsuperscript{\rm 2}Monash University,  \\
    \textsuperscript{\rm 3}Zhejiang University, \\
    \textsuperscript{\rm 4}University of Chinese Academy of Sciences,  \\
    \textsuperscript{\rm 5}Tsinghua University \\

}

\usepackage{bibentry}

\begin{document}

\maketitle


\input{section/Abstract}
\input{section/Introduction}
\input{section/related}
\input{section/Method}

\input{section/Exp}
\input{section/con}

\section*{Acknowledgments}
This work was supported in part by the National Natural Science Foundation of China under Grants 62471168, 62422204, and 61802100, 62372147, U21B2040 and in part by the Zhejiang Provincial Natural Science Foundation of China under Grant LDT23F02025F02 and LY21F020019.


\bibliography{aaai2026}

\end{document}

%% file: section/Abstract.tex
\begin{abstract}

Recent advances in instruction-based image editing have shown remarkable progress. However, existing methods remain limited to relatively simple editing operations, hindering real-world applications that require complex and compositional instructions. In this work, we address these limitations from the perspectives of \textit{architectural design}, \textit{data}, and \textit{evaluation protocols}. Specifically, we identify two key challenges in current models: insufficient instruction compliance and background inconsistency. To this end, we propose \textbf{MCIE-E1}, a Multimodal Large Language Model–Driven Complex Instruction Image Editing method that integrates two key modules: a \textit{spatial-aware cross-attention} module and a \textit{background-consistent cross-attention} module. The former enhances instruction-following capability by explicitly aligning semantic instructions with spatial regions through spatial guidance during the denoising process, while the latter preserves features in unedited regions to maintain background consistency. To enable effective training, we construct a dedicated data pipeline to mitigate the scarcity of complex instruction-based image editing datasets, combining fine-grained automatic filtering via a powerful MLLM with rigorous human validation. Finally, to comprehensively evaluate complex instruction-based image editing, we introduce \textbf{CIE-Bench}, a new benchmark with two new evaluation metrics. Experimental results on CIE-Bench demonstrate that MCIE-E1 consistently outperforms previous state-of-the-art methods in both quantitative and qualitative assessments, achieving a 23.96\% improvement in instruction compliance.

\end{abstract}

%% file: section/Introduction.tex
\section{Introduction}
\begin{figure}[t]
\includegraphics[width=\columnwidth]{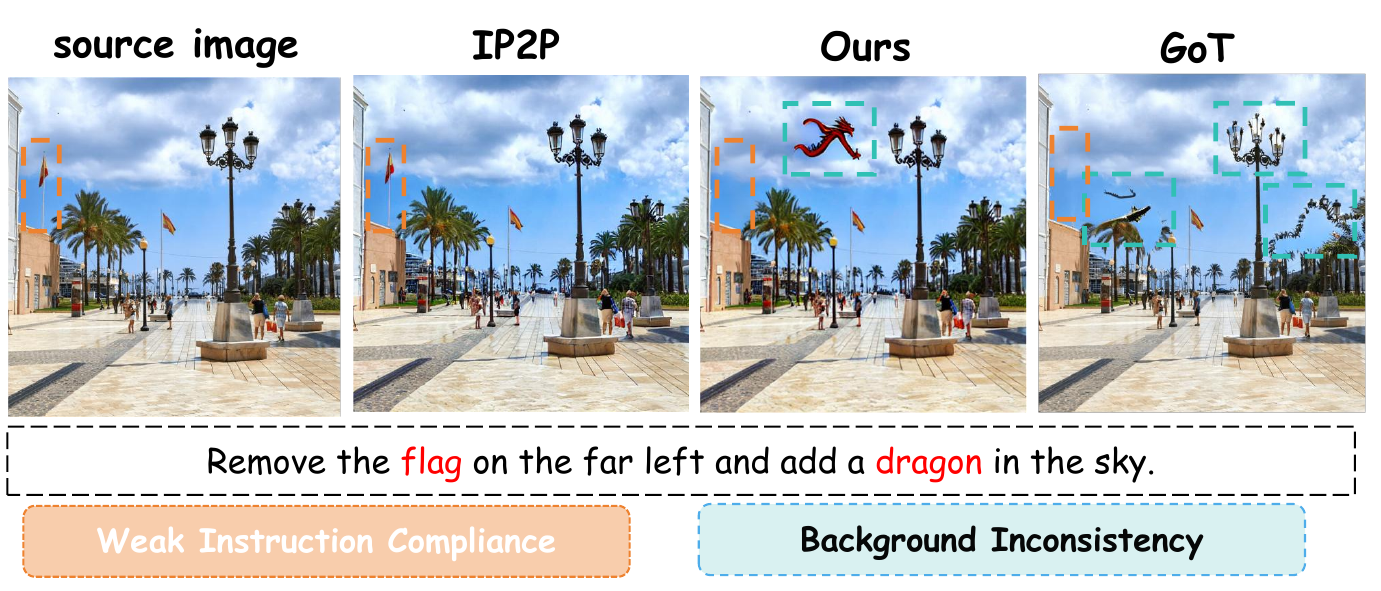} 
\caption{We propose MCIE-E1 to address the challenges of weak instruction compliance and background consistency in complex instruction-based image editing. }
\label{motivation}
\end{figure}


Diffusion models have achieved notable progress in instruction-based image editing, attracting substantial attention from both academia and industry. These tasks range from simple edits \cite{brooks2023instructpix2pix, yu2025anyedit, sheynin2024emu} that modify a single element within an image to complex manipulations \cite{guo2024focus} requiring the simultaneous editing of multiple regions. Although existing diffusion-based models \cite{zhao2024ultraedit, yu2025anyedit} perform well on simple editing instructions, they often struggle with detailed and compositional modifications. This gap between simple and complex editing capabilities in current models raises a fundamental question: how can we advance instruction-based image editing toward reliable and controllable complex editing?

Prior work \cite{yu2025anyedit,hertz2022prompt} has explored instruction-based image editing from different perspectives. One line of research leverages CLIP~\cite{ramesh2022hierarchical} to extract intended modifications from user instructions. Another line of research \cite{huang2024smartedit,fang2025got} integrates multimodal large language models (MLLMs) into diffusion models to enhance the understanding of fine-grained editing instructions. As shown in Fig.~\ref{motivation}, when users provide complex instructions, existing approaches~\cite{brooks2023instructpix2pix,fang2025got,ma2024followpose} often either overlook sub-instructions, leading to suboptimal instruction-following, or introduce unintended changes, compromising background consistency.

\begin{figure*}
\centering
\includegraphics[width=0.96\linewidth]{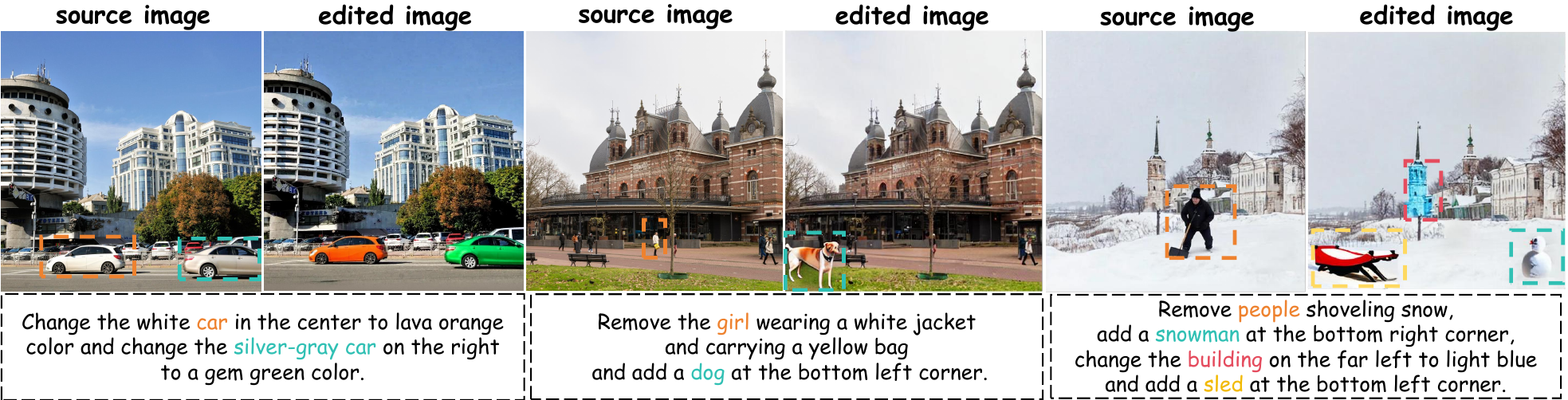}
  \caption{Visual results of MCIE-E1. Our method effectively performs complex instruction-based image editing with accurate and consistent outputs.}
  \label{fig:teaser}
\vspace{-0.5em}
\end{figure*}
Current methods still face challenges in  complex instruction-based image editing, primarily for the following reasons:
(1) \emph{\textbf{
Lack of high-quality data for complex instruction editing}}. Existing data creation pipelines \cite{zhang2023magicbrush, brooks2023instructpix2pix, yu2025anyedit, zhao2024ultraedit,bai2024humanedit} mainly focus on simple instructions and low-resolution images, and often lack fine-grained, multi-stage post-processing, limiting the model’s ability to perform instruction-based image editing.
(2) \emph{\textbf{Inadequate instruction-region alignment}}. Although recent methods \cite{brooks2023instructpix2pix, fu2023guiding, yu2025anyedit, huang2024smartedit, fang2025got,ma2025followyourmotion} have made notable advances, they still struggle to precisely localize editing regions for complex instructions, leading to incomplete or inaccurate edits. This misalignment also affects the model’s ability to preserve background consistency and accurately follow instructions, further reducing output reliability.
(3) \emph{\textbf{Limited benchmarking protocols}}. Current evaluation frameworks \cite{ma2024i2ebench, sheynin2024emu, wang2023imagen, basu2023editval, kawar2023imagic,zhang2026well} often fall short in comprehensively assessing a model’s ability to follow complex instructions and maintain background consistency, which hinders the effective evaluation of complex instruction-based image editing.

\begin{figure*}[t]
\centering
\includegraphics[width=1.0\textwidth,page=1]{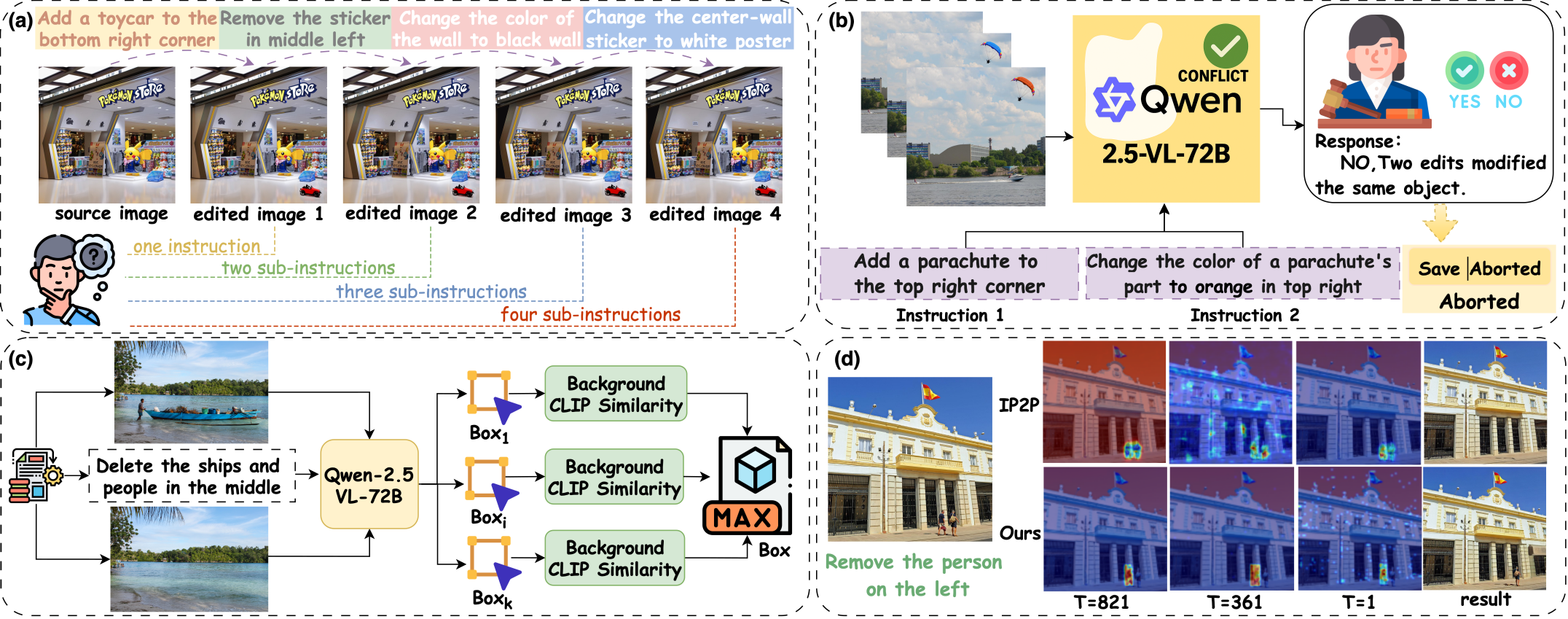} 
\caption{
(a) shows how multi-turn editing sequences are expanded into multiple complex instruction editing instances. (b) shows the use of Qwen2.5-VL-72B to detect instruction conflicts. 
(c) illustrates the generation and selection of bounding boxes. (d) compares attention maps and editing results for IP2P and our method during the denoising process.
}
\label{fig1}
\end{figure*}

To address these challenges, we propose a unified framework that combines: 
(1)\emph{\textbf{ A large-scale and high-quality dataset.}} We introduce the MCIE dataset, the first dataset tailored for complex instruction image editing. First, we aggregate multi-round edits into complex instructions, which may potentially introduce semantic conflicts. As illustrated in Fig.~\ref{fig1}(b), we use Qwen2.5-VL-72B \cite{bai2025qwen2} to detect conflicts in the augmented dataset. Additionally, we generate bounding boxes to provide spatial guidance for complex instructions. Finally, human experts meticulously filter the data along three dimensions: instruction consistency, image quality, and editing scenario complexity.
(2)\emph{\textbf{ A novel image editing model for complex instruction.}} We propose MCIE-E1, a model for complex instruction-based image editing. It leverages MLLMs to disentangle complex instructions into sub-instructions with their corresponding editing regions. Specifically, we introduce spatial-aware cross-attention, which processes sub-instructions independently and injects spatial guidance to enhance instruction following. To preserve background consistency, we propose background-consistent cross-attention, which utilizes pixel-level visual features to maintain coherence in unedited regions.
(3)\emph{\textbf{ A general benchmark for evaluating complex instruction-based image editing.}} We construct CIE-Bench, which includes 400 carefully crafted evaluation sets, and we introduce two metrics to assess the model’s ability to follow instructions and maintain background consistency.

By leveraging MLLMs to decompose complex instructions and incorporating a semantically and spatially aware diffusion model, MCIE-E1 achieves superior instruction compliance and background consistency, as shown in Fig.~\ref{fig:teaser}. Extensive experimental results demonstrate that our method outperforms existing approaches on complex instruction-based image editing. 

Our main contributions are summarized as follows:

\begin{itemize}
    \item We tackle the novel task of complex instruction-based image editing and present the MCIE dataset, which comprises 90k samples carefully filtered by human experts along multiple dimensions. 
    \item To address the challenge of complex instruction-based image editing, we propose MCIE-E1, a novel model with two key components: a spatial-aware cross-attention module for enhancing instruction following and a background-consistent cross-attention module for preserving unedited areas.    
    \item To validate the effectiveness of our method, we construct CIE-Bench and introduce two evaluation metrics, Instruction Compliance and Background Consistency, on which MCIE-E1 demonstrates clear superiority. 
\end{itemize}

%% file: section/related.tex
\section{Related Work}
\subsection{Instruction-Based Image Editing}

IP2P \cite{brooks2023instructpix2pix} is a pioneering work that employs instructions to edit images by fine-tuning a diffusion model on a large-scale dataset. InstructDiffusion \cite{geng2024instructdiffusion} further generalizes instruction-based image editing to conventional vision tasks. AnyEdit \cite{yu2025anyedit} introduces a Mixture-of-Experts mechanism to address diverse editing scenarios. Despite these advancements, most existing methods rely on CLIP \cite{ramesh2022hierarchical} or T5 \cite{raffel2020exploring}, which limits their ability to interpret fine-grained instructions. To mitigate this limitation, SmartEdit \cite{huang2024smartedit} and MGIE \cite{fu2023guiding} integrate MLLMs into diffusion-based frameworks to enhance instruction comprehension. However, these approaches still exhibit limited performance in handling complex instructions. The most relevant work to our study is FOI \cite{guo2024focus}, which leverages the cross-attention mechanism in U-Net to localize editing regions. Nevertheless, due to the limited comprehension capability of CLIP, FOI often fails to generate accurate masks for the edited areas. Our key insight is that general diffusion models struggle to execute complex instructions, making it essential to decompose them into sub-instructions with corresponding spatial regions. Furthermore, we incorporate spatial information for each sub-instruction to improve instruction-following performance and integrate fine-grained visual features to maintain background consistency.

\subsection{Datasets for Instruction Editing}


In Tab.~\ref{tab:dataset-comparison}, we compare several popular instruction-based image editing datasets~\cite{brooks2023instructpix2pix,zhang2023magicbrush,yu2025anyedit,wei2024omniedit}. For instance, IP2P uses P2P~\cite{hertz2022prompt} to generate suboptimal image pairs, which limits its applicability in real-world editing scenarios. MagicBrush~\cite{zhang2023magicbrush} improves real-world applicability through 10,000 high-quality human annotations. Both OmniEdit~\cite{wei2024omniedit} and AnyEdit~\cite{yu2025anyedit} leverage expert models to generate diverse image editing tasks. UltraEdit~\cite{zhao2024ultraedit} scales up data generation by incorporating both free-form and region-based samples. SEED-Data-Edit~\cite{ge2024seed} provides single-turn and multi-turn editing data, covering a broader range of real-world images. However, these datasets still suffer from limited data quality and overly simple instructions. In contrast, our MCIE dataset is specifically tailored for complex instruction-based image editing.

%% file: section/Method.tex
\section{Method}
\begin{table*}[ht]
\centering

 \resizebox{\linewidth}{!}{
\begin{tabular}{lccccc|ccc}
\toprule
\textbf{Dataset} & \textbf{Size} & \textbf{Res. (px)$\uparrow$} & \textbf{DeQA-Score$\uparrow$} & \textbf{Q-Insight$\uparrow$} & \textbf{Q-align$\uparrow$}& \textbf{CE} & \textbf{SI} & \textbf{HF} \\
\midrule
InstructPix2Pix~\cite{brooks2023instructpix2pix}   & 313K  & 512      & 3.38 & 3.61 & 3.67  & \textcolor{red}{\ding{55}}  & \textcolor{red}{\ding{55}} & \textcolor{red}{\ding{55}}  \\
MagicBrush~\cite{zhang2023magicbrush} & 10K & 500      & 3.31 & 3.32 & 3.35  & \textcolor{red}{\ding{55}}     & \textcolor{green!60!black}{\ding{51}} & \textcolor{green!60!black}{\ding{51}} \\

OmniEdit~\cite{wei2024omniedit}   & 1.2M  & $\geq$768  & 3.44 & 3.69 & 3.83  & \textcolor{red}{\ding{55}}  & \textcolor{red}{\ding{55}} & \textcolor{red}{\ding{55}}  \\
AnyEdit~\cite{yu2025anyedit}    & 2.5M & 512      & 3.65 & 3.70 & 3.92 & \textcolor{red}{\ding{55}}   & \textcolor{red}{\ding{55}} & \textcolor{red}{\ding{55}}  \\
UltraEdit~\cite{zhao2024ultraedit}  & 4M   & 512  & 3.65 & 3.67 & 3.75  & \textcolor{red}{\ding{55}}   & \textcolor{green!60!black}{\ding{51}} & \textcolor{red}{\ding{55}}  \\
SEED-Data-Edit~\cite{ge2024seed}   & 3.7M  & $\geq$768  & 3.59 & 3.73 & 3.80&  \textcolor{red}{\ding{55}}  & \textcolor{red}{\ding{55}} & \textcolor{red}{\ding{55}}  \\

\textbf{MCIE dataset (ours)} & 90K  & $\geq$1024 & \textbf{3.90} & \textbf{3.95} & \textbf{4.27} & \textcolor{green!60!black}{\ding{51}} & \textcolor{green!60!black}{\ding{51}} & \textcolor{green!60!black}{\ding{51}}  \\
\bottomrule
\end{tabular}}
\caption{Comparison of existing datasets and the proposed MCIE dataset. \textbf{CE} denotes Complex Editing, \textbf{SI} denotes Spatial Information, and \textbf{HF} denotes  Human Filter. \textbf{DeQA-Score} \cite{you2025teaching}, \textbf{Q-Insight} \cite{li2025q}, and \textbf{Q-align} \cite{wu2023q} are used to assess image quality. }
\label{tab:dataset-comparison}
\end{table*}
We present a high-fidelity dataset and a state-of-the-art model for complex instruction-based image editing. Section~\ref{Dataset} details the data creation pipeline, including data augmentation, data filtering, spatial information generation, and post-processing. Section~\ref{Method} introduces MCIE-E1, an advanced image editing model trained on a subset of OmniEdit-GoT \cite{fang2025got} and the MCIE dataset.

\subsection{MCIE Dataset}
\label{Dataset}

Existing datasets~\cite{zhang2023magicbrush, yu2025anyedit,chen2025opengpt} primarily support simple instructions, limiting their effectiveness for complex instruction-based image editing. Moreover, complex instructions often require auxiliary inputs such as bounding boxes or masks to provide spatial guidance. To address these limitations, we construct a large-scale, high-quality dataset by leveraging the understanding and grounding capabilities of advanced MLLMs. Our MCIE dataset includes meticulously designed semantic–spatial information for complex instruction-based image editing, with each sample containing an instruction, bounding boxes, and corresponding images. Finally, to ensure data quality, experts evaluate the dataset based on three criteria: \emph{instruction consistency}, \emph{image quality}, and \emph{editing scenario complexity}.

\noindent\textbf{Data Preparation and Augmentation.} As the foundation of our study, we adopt 20K multi-turn editing instances from the SEED-Data-Edit \cite{ge2024seed} dataset as our primary corpus, as it offers more diverse instructions and more accurate editing results than other datasets. However, the dataset contains only 21K multi-round samples, limiting the generalization capability of models. To address this, we propose an efficient data augmentation approach. As illustrated in Fig.~\ref{fig1}(a), a four-turn editing instance can be expanded into \textbf{three} two-instruction, \textbf{two} three-instruction, and \textbf{one} four-instruction editing samples. Nevertheless, this approach inherits the limitations of multi-turn editing, which may involve modifying the same object multiple times.

\noindent\textbf{Data Filtering.} As illustrated in Fig. \ref{fig1}(b), we provide the original image, the intermediate results after each edit, and the corresponding instruction for each turn to facilitate conflict detection. The MLLM must comprehend the editing instructions for each turn along with their associated regions, enabling it to reason about whether the same object undergoes multiple modifications. Leveraging the strong multi-image understanding capabilities of Qwen2.5-VL-72B \cite{bai2025qwen2}, we employ carefully designed in-context prompts to evaluate whether converting multi-turn edits into complex instruction edits introduces any conflicts.

\noindent\textbf{Post Processing.} Since coarse filtering has already been performed during data preparation using MLLMs, we further employ 20 human experts to conduct a fine-grained evaluation across three key aspects: instruction consistency, image quality, and editing scenario complexity. Instruction consistency is assessed with two options, i.e., ``Yes'' or ``No'', ensuring that the editing instructions are conflict-free. For image quality and editing scenario complexity, we adopt a 1–5 scoring system with detailed criteria, retaining only samples with scores above 3.

\noindent\textbf{Spatial Information Generation.} Spatial information is critical in image editing, significantly affecting instruction-following performance. We decompose multi-turn editing into single-turn instructions. As depicted in Fig.~\ref{fig1}(c), leveraging the strong grounding capabilities of Qwen2.5-VL-72B, we generate $k$ candidate bounding boxes for each instruction and select the one that best aligns with the edited region. Because the bounding box primarily covers the areas to be edited, the remaining regions of the source and target images should remain consistent. Accordingly, we compute the CLIP score between the source and target images outside the bounding box to select the most accurate box.
\begin{equation}
    i^* = \underset{i \in \{1,2,\dots,k\}}{\arg\max}\ \text{CLIP}\bigl(I_{\text{src}} \cdot (1-M_i), I_{\text{tar}} \cdot (1-M_i)\bigr)
\end{equation}
where $I_{\text{src}}$ and $I_{\text{tar}}$ denote the source and target images, respectively. $M_i$ denote the mask associated with the $i$-th candidate bounding box.

The creation of the entire dataset requires substantial computational resources, specifically utilizing 32 H20 GPUs 
over a week. This robust computational setup ensures the precision and reliability of instructions and 
bounding boxes, thereby providing a solid and essential foundation for training models on complex instruction image editing.
\begin{figure}[t]
\centering
\includegraphics[width=\linewidth]{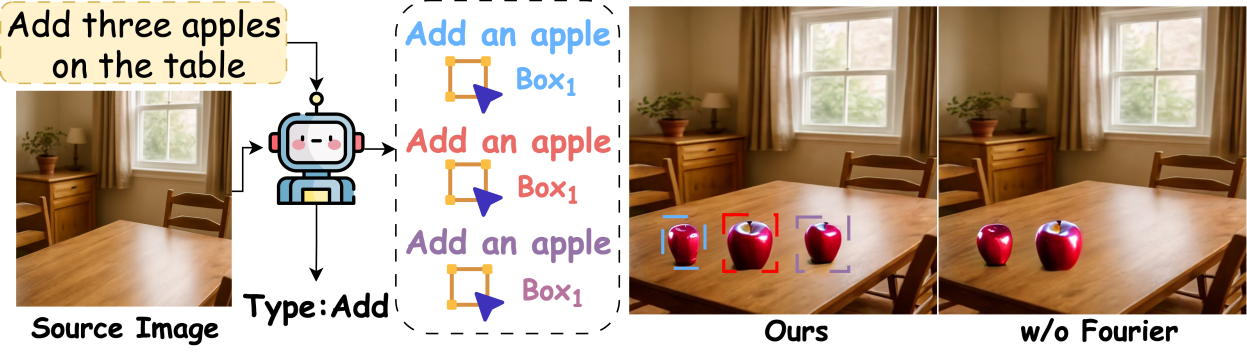} 
\caption{Comparison of editing results for sub-instruction bounding boxes with and without the Fourier transform. }
\label{fouier}
\end{figure}

\subsection{Complex Instruction Image Editing Design}
\label{Method}
\begin{figure*}[t]
\centering
\includegraphics[width=1.0\linewidth]{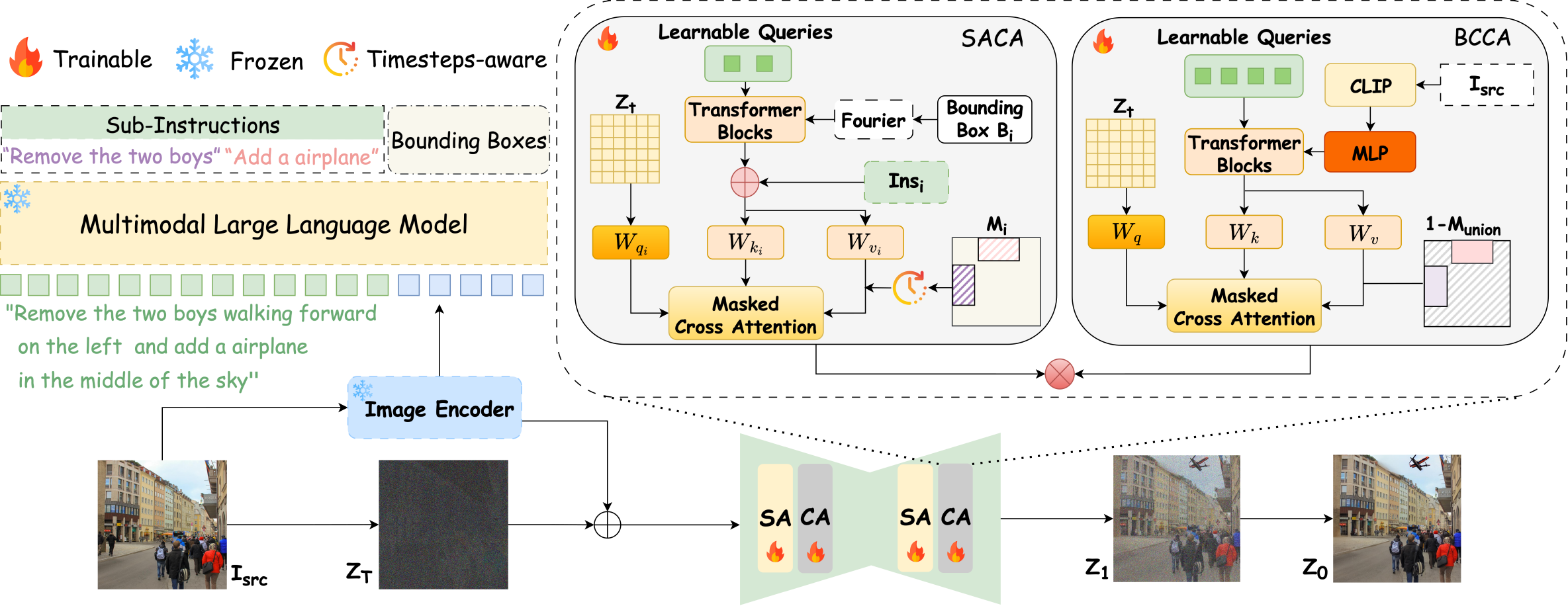} 
\caption{The overall framework of MCIE-E1. It employs an MLLM for instruction decomposition and guiding the diffusion model through two key modules: SACA for enhancing instruction following and BCCA for preserving non-edited regions.}
\label{fig4}
\end{figure*}

Given a source image and a complex instruction 
\( I = \{I_1, I_2, ..., I_m\} \), where each sub-instruction $I_i$ is associated  with operation type \(op_i \in \{\text{ADD}, \text{REMOVE}, \text{CHANGE}\}\). 
As shown in Fig.~\ref{fig4}, MCIE-E1 integrates an MLLM to decompose complex instructions and a diffusion model comprising two key components: spatial-aware cross-attention for enhancing instruction following and background-consistent cross-attention for maintaining background consistency.

\noindent\textbf{Instruction Decomposition.} Recently, MLLMs have shown remarkable capabilities in understanding, reasoning, and few-shot generalization. We adopt an in-context learning strategy to enable accurate instruction decomposition and region generation. Since diffusion models inherently struggle with counting, we further decompose quantity-related instructions to enhance the model’s instruction-following capability. The decomposition process is as follows:
\begin{equation}
    \mathcal{D}: I \rightarrow \{(I_1, B_1), (I_2, B_2), ..., (I_m, B_m)\}
\end{equation}
\noindent where $B_i$ denotes the bounding box corresponding to the editing region specified by sub-instruction $I_i$.


The prompt consists of three integrated components: a carefully designed guideline specifying the assistant’s role and decomposition format, a representative example illustrating the decomposition of a complex instruction into sub-instructions, and a test instance containing a complex instruction and a source image.




\noindent\textbf{Spatial-Aware Cross-Attention.} Previous studies \cite{brooks2023instructpix2pix,huang2024smartedit,fang2025got} typically encode entire instructions using a CLIP \cite{radford2021learning} model, which causes unintended interference among sub-instructions. To mitigate this, we introduce an instruction-wise encoding strategy, where each sub-instruction is encoded independently and in parallel. The instruction-wise encoding strategy is defined as:
\begin{equation}
    \text{Ins}_i = \text{CLIP}(\text{I}_i)
\end{equation}
\begin{equation}
\text{T} = [\text{Ins}_1, \dots, \text{Ins}_m]
\end{equation}
\noindent where \([ \cdot ]\) denotes concatenation.
 



For complex image editing, spatial information is crucial for guiding the model to execute modifications accurately. As shown in Fig.~\ref{fig1}(d), IP2P~\cite{brooks2023instructpix2pix} often produces imprecise edits. In contrast, our model incorporates rich spatial cues, enabling more controllable and precise editing. We address this by employing masked cross-attention to associate each sub-instruction with its corresponding region explicitly. However, this approach may still encounter failures when processing semantically similar instructions.
As illustrated in Fig.~\ref{fouier}, when the instruction is to add three apples on the table, it is decomposed into three semantically similar sub-instructions. Due to their similarity, the sub-instructions focus on the same region, causing the model to execute only part of the intended edits. To this end, we propose a Spatial-Aware Cross-Attention (SACA) module that encodes bounding boxes via Fourier mapping and extracts features using learnable queries, enabling the model to learn distinct spatial embeddings for each sub-instruction. This process is formalized as follows:
\begin{equation}
    \mathcal{F}(\mathbf{B}_i) = \big[ \sin(2\pi \mathbf{B}_i \mathbf{f}_k), \cos(2\pi \mathbf{B}_i \mathbf{f}_k) \big]
\end{equation}
\begin{equation}
     \text{C}_i = [\text{T}_i,Q_{\beta}(\mathcal{F}(\mathbf{B}_i))] \\
\end{equation} 
\begin{equation}
    \text{FG}_t = \text{Softmax} \Big( \frac{Q K_i^\top \odot M_i}{\sqrt{d}} \Big) V_i
\end{equation}
\noindent
where $\mathbf{f}_k$ are frequency bands, $Q_{\beta}$ denotes $L_1$ Transformer blocks, $\odot$ represents element-wise multiplication,  $\text{FG}_t$ are foreground features,
\( Q = z_tW_q \), \( K_i = C_i W_{k_i} \), \( V_i = C_i W_{v_i} \), $z_t$ is the noisy latent, and $d$ is the feature dimension.


Prior studies \cite{hu2024ella,zhang2023prospect} have shown that diffusion models in text-to-image generation primarily attend to spatial layouts and global structures in the early stages of denoising, gradually shifting focus to fine-grained features. We observe that instruction-based image editing exhibits a similar progression, focusing on different types of information throughout the denoising process. Inspired by this observation, we adopt a timestep-aware control schedule to modulate the strength of spatial guidance.
Specifically, we define a timestep-aware mask: 
\begin{equation}
    M_i^{'} = M_i * \exp\left(-\frac{t}{T}\right)
\end{equation}
\noindent where \( t \) denotes the current denoising timestep, and \( T \) is the total number of denoising steps.

\begin{table*}[t]
\centering

\resizebox{0.82\textwidth}{!}{ 
\begin{tabular}{lcccccc|cc}
\toprule
\textbf{Method} &
\multicolumn{6}{c|}{\textbf{Automatic Metrics}} &
\multicolumn{2}{c}{\textbf{User Study}} \\
\cmidrule(r){2-7} \cmidrule(l){8-9}
& CLIP-I$\uparrow$ & DINO-I$\uparrow$ & L1$\downarrow$ & L2$\downarrow$ & IC$\uparrow$ & BC$\uparrow$ 
& Ins Following$\uparrow$ & BG Consist.$\uparrow$ \\
\midrule
IP2P & 0.8784 & 0.9691 & 0.0710 & 0.1032 & 0.4922 & 0.6270 & 1.00\% & 2.00\% \\
MagicBrush & 0.8870 & 0.9813 & \textcolor{blue}{\textbf{0.0476}} & \textcolor{blue}{\textbf{0.0992}} & 0.5246 & 0.7608 & 0.67\% & 1.33\% \\
UltraEdit & 0.7820 & 0.9580 & 0.0779 & 0.1209 & 0.5495 & 0.6485 & \textcolor{blue}{\textbf{6.67\%}} & 6.00\% \\
AnyEdit & 0.8847 & 0.9662 & 0.0701 & 0.1168 & 0.5335 & 0.7659 & 0.67\% & 2.33\% \\
InsDiff & 0.8815 & 0.9634 & 0.0810 & 0.1283 & 0.4812 & 0.7465 & 1.00\% & 1.67\% \\
MGIE & 0.7679 & 0.9185 & 0.1407 & 0.2005 & 0.3988 & 0.6205 & 0.33\% & 0.33\% \\
SmartEdit & 0.8836 & 0.9752 & 0.0629 & 0.1042 & 0.5416 & 0.7419 & 3.00\% & 4.33\% \\
GoT & 0.8856 & 0.9799 & 0.0549 & 0.1078 & 0.6491 & 0.7389 &  \textcolor{blue}{\textbf{6.67\%}} & \textcolor{blue}{\textbf{9.67\%}} \\
MCIE-E1$\dagger$ & \textcolor{red}{\textbf{0.9200}} & \textcolor{red}{\textbf{0.9863}} & \textcolor{red}{\textbf{0.0456}} & \textcolor{red}{\textbf{0.0864}} & \textcolor{blue}{\textbf{0.6597}} & \textcolor{blue}{\textbf{0.7784}} & -- & -- \\ 
\textbf{MCIE-E1} & \textcolor{blue}{\textbf{0.8899}} & \textcolor{blue}{\textbf{0.9833}} & 0.0489 & 0.1029 & \textcolor{red}{\textbf{0.8046}} & \textcolor{red}{\textbf{0.7831}} & \textcolor{red}{\textbf{80.00\%}} & \textcolor{red}{\textbf{72.33\%}} \\
\bottomrule
\end{tabular}
} 
\caption{Quantitative results on CIE-Bench. $\uparrow$ indicates higher is better, and $\downarrow$ indicates lower is better. The best result in each column is highlighted in \textcolor{red}{\textbf{red}}, and the second best in \textcolor{blue}{\textbf{blue}}. $\textbf{MCIE-E1}\dagger$ denotes a variant trained only in the first stage.}
\label{22}
\end{table*}

\noindent\textbf{Background-Consistent Cross-Attention.} 
Complex editing instructions often involve multiple region-specific operations, increasing the risk of background inconsistency and potentially compromising the visual coherence of the edited image. To address this, we introduce the Background-Consistent Cross-Attention (BCCA) module. As shown in Fig.~\ref{fig4}, we first extract pixel-level visual features from the source image using the CLIP~\cite{radford2021learning} model. These features are then refined through an MLP followed by $L_2$ Transformer blocks, with a set of learnable queries capturing enhanced representations. Finally, the learnable queries interact via masked cross-attention, ensuring high background consistency throughout the denoising process.
\begin{equation}
    f = Q_{\beta}\big( \text{MLP}(\text{CLIP}(I_{\text{src}})) \big)
\end{equation}
\begin{equation}
    \text{BG}_t = \text{Softmax}\Bigg( \frac{Q K_t^\top \odot (1 - M_{\text{union}})}{\sqrt{d}} \Bigg) V_t
\end{equation}
\noindent
where 
\(
K_t = W_k f, 
V_t = W_v f, 
M_{\text{union}} = \bigcup_{i=1}^{m} M_i
\), $\text{BG}_t$ are background features.




Following IP-Adapter \cite{ye2023ip}, we decouple the cross-attention layers into two key modules: SACA, which enhances interactions between foreground features and complex instructions, and BCCA, which preserves the consistency of unedited background regions. Specifically, given the foreground features $\text{FG}_t$ and background features $\text{BG}_t$, they are fused as:
\begin{equation}
z_t^{\prime} = \lambda \cdot \text{FG}_t + (1 - \lambda) \cdot \text{BG}_t
\end{equation}
\noindent where $\lambda \in [0,1]$ is a weighting factor.


\noindent\textbf{Training Procedure.} 
We adopt a two-phase training strategy. We first train the model for 20,000 steps on a subset of the OmniEdit-GoT \cite{fang2025got} dataset. Subsequently, we fine-tune the model for an additional 10,000 steps on the MCIE dataset to better adapt to complex and fine-grained editing instructions. The optimization process for \( \epsilon_\theta \) can be expressed in two stages as follows:
\begin{equation}
\mathcal{L} = \mathbb{E}_{\mathcal{E}(y), \mathcal{E}(x), c, \epsilon, t} 
\left\| \epsilon - \epsilon_{\theta}\left(\text{Cat}(z_t, \mathcal{E}(x)), B, y, t\right) \right\|_2^2
\end{equation} 
\noindent where  $x$ is the source image, $y$ is the target image, \( \mathcal{E}(x) \) is the encoded image latent, \( z_t \) is the noisy latent,  \( \epsilon_\theta \) is the denoising network and Cat denotes concatenation along the channel dimension.



%% file: section/Exp.tex
\section{Experiment}
\subsection{CIE-Bench}
We introduce CIE-Bench, a new benchmark to evaluate a model’s ability to perform complex instruction-based image editing.
Specifically, we select 400 high-quality images 
from the SEED-data-Edit dataset, 
covering a variety of realistic scenarios. Additionally, we employ GPT-4o~\cite{hurst2024gpt} to 
generate complex instructions, each containing 1 to 4 
sub-instructions involving any combination of three operation types: 
\emph{Add}, \emph{Change}, and \emph{Remove}. To ensure accurate evaluation, 
human experts manually annotate the corresponding editing regions based on the instructions, and each sample undergoes manual filtering.

\noindent\textbf{Evaluation Metrics.}
To quantitatively assess editing performance, we use CLIP \cite{radford2021learning}, DINOv2 \cite{oquab2023dinov2} image similarity, L1 distance, and L2 distance.   


Additionally, we introduce two new evaluation metrics:
\begin{itemize}
\item \textbf{Instruction Compliance (IC)}: GPT-4o  is employed to assess whether the specified modifications align with the given instructions, rated on a scale of 1–10.
\item \textbf{Background Consistency (BC)}: GPT-4o scores the consistency of the unedited regions on a scale of 1–5.
\end{itemize}

The instruction compliance and background consistency scores are then normalized to the range $[0, 1]$.


\subsection{Experimental Settings}
\begin{figure*}[htbp]
\centering
\includegraphics[width=1.0\textwidth]{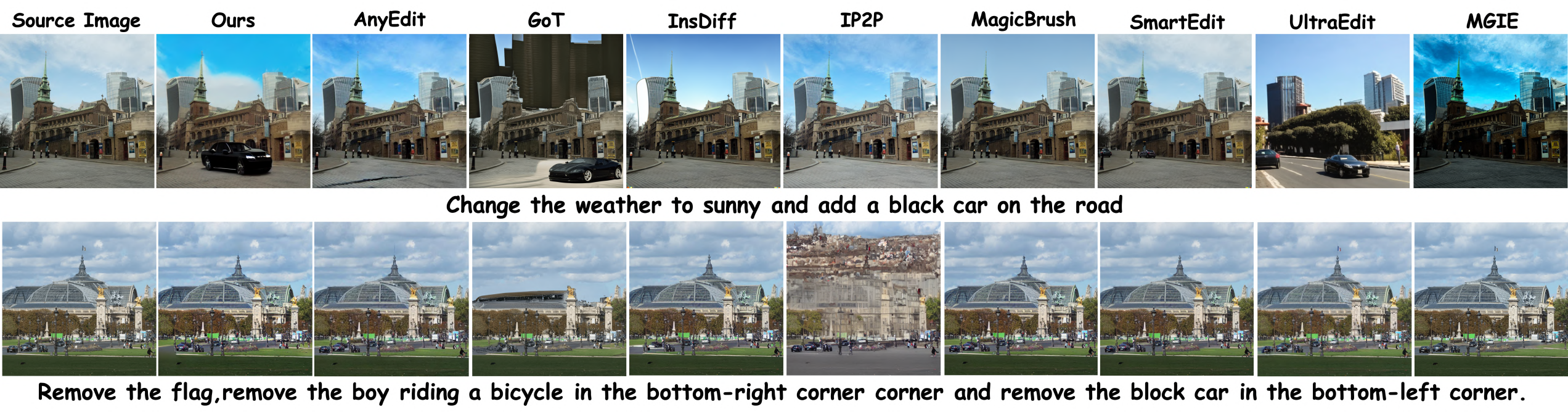} 
\caption{Qualitative results on CIE-bench. Our MCIE-E1 model demonstrates superior performance in terms of instruction compliance and background consistency.}

\label{fig5}
\end{figure*}
\noindent\textbf{Baseline Models.}
We evaluate our MCIE-E1 model against state-of-the-art image editing methods, including  IP2P \cite{brooks2023instructpix2pix}, MagicBrush \cite{brooks2023instructpix2pix}, InstructDiffusion(InsDiff) \cite{geng2024instructdiffusion}, AnyEdit \cite{yu2025anyedit}
, UltraEdit \cite{zhao2024ultraedit}, MGIE \cite{fu2023guiding}, SmartEdit \cite{huang2024smartedit} and GoT \cite{fang2025got}. 

\noindent\textbf{Implementation Details.} We implement MCIE-E1 on Stable Diffusion 1.5~\cite{rombach2022high}, with parameters initialized from IP2P \cite{brooks2023instructpix2pix}. Low-Rank Adaptation (LoRA) \cite{hu2022lora} is employed to efficiently update only the self-attention layers, while the SACA and BCCA modules are fully optimized within the diffusion model. The SACA module consists of two Transformer blocks and two learnable queries, whereas the BCCA module comprises four Transformer blocks and 16 learnable queries. All experiments use the Euler Ancestral sampler with 20 denoising steps.


\subsection{Main Results}
\noindent\textbf{Quantitative Evaluation.} Tab.~\ref{22} presents the quantitative comparison of our method with baseline methods on CIE-Bench. Our method consistently outperforms all baselines across all automatic evaluation metrics, demonstrating its effectiveness in complex instruction-based image editing. Tab.~\ref{magicbrush} shows the quantitative comparison on MagicBrush for simple instruction-based image editing. Similarly, our method achieves superior performance across all metrics, indicating its advantage over baselines in simpler editing tasks as well.


\noindent \textbf{Qualitative Evaluation.} Fig.~\ref{fig5} presents qualitative comparisons on CIE-Bench. As observed, baseline methods often struggle to follow specific editing instructions. For instance, IP2P~\cite{brooks2023instructpix2pix} and MagicBrush~\cite{zhang2023magicbrush} fail to add a black car on the road, while SmartEdit~\cite{huang2024smartedit} cannot change the weather to sunny. Moreover, they struggle to maintain background consistency. For example, GoT \cite{fang2025got} modifies buildings across the entire image when instructed to make the weather sunny, leading to severe distortions. In contrast, our method performs precise edits while preserving background consistency.

\begin{table}[htbp]
\centering

\label{tab:comparison}
\begin{tabular}{lcccc}
\toprule
Method & L1$\downarrow$ & L2$\uparrow$ & CLIP-I$\uparrow$ & DINO-I$\uparrow$  \\
\midrule
IP2P & 0.1059 & 0.0328 & 0.8674 & 0.7617 \\
MagicBrush & 0.0765 & 0.0270 & 0.8951 & 0.8360 \\
AnyEdit & 0.0968 & 0.0383 & 0.8689 & 0.7532  \\
InsDiff & 0.0905 & 0.0363 & 0.8934 & 0.8045  \\
MGIE & 0.0847 & 0.0324 & 0.9033 & 0.8292  \\
SmartEdit & 0.0672 & 0.0284 & 0.9017 & 0.8360 \\
GoT &  \textcolor{blue}{\textbf{0.0645}} &  \textcolor{blue}{\textbf{0.0237}} & 0.9047 & 0.8400 \\
\midrule
MCIE-E1$\dagger$ &  \textcolor{red}{\textbf{0.0641}} &  \textcolor{red}{\textbf{0.0228}} &  \textcolor{red}{\textbf{0.9143}} & \textcolor{red}{\textbf{0.8497}} \\
MCIE-E1 & 0.0669 & 0.0246 &  \textcolor{blue}{\textbf{0.9122}} &  \textcolor{blue}{\textbf{ 0.8466}} \\
\bottomrule
\end{tabular}
\caption{ Quantitative results on MagicBrush. 
}
\label{magicbrush}
\end{table}


\noindent \textbf{User Study.} We conducted a user study on complex instruction-based image editing with 50 participants, each evaluating 100 images. The study assessed user preferences across two key aspects: \emph{instruction following} and \emph{background consistency}. Participants were asked to select the method that best adhered to the given instructions and best preserved non-edited regions. As shown in Tab.~\ref{22}, our method was consistently preferred over baseline methods, with 80\% of participants indicating that its instruction-following capability surpassed that of the baselines.
\subsection{Ablation Study} We conduct an ablation study to evaluate the contribution of each component in the proposed method on CIE-bench, including the SACA and BCCA modules. In addition, we examine the impact of different MLLMs.

\noindent\textbf{Impact of Spatial-Aware Cross-Attention.} As shown in Tab.~\ref{table3}, removing the SACA module markedly degrades the model’s ability to follow sub-instructions, whereas its inclusion substantially enhances instruction-following performance.
 
\noindent\textbf{Impact of Background-Consistent Cross-Attention.} As shown in Tab.~\ref{table3}, integrating the BCCA module yields notable gains in CLIP-I, DINO-I, and BC metrics, highlighting the effectiveness of its pixel-level visual encoding in maintaining background consistency.
\begin{table}[ht]
\centering
\resizebox{\linewidth}{!}{  
\input{table/Ablation_module}
}
\caption{Quantitative ablation results of SACA and BCCA modules.}
\label{table3}
\end{table}

\noindent\textbf{Impact of MLLMs.} We further perform ablation studies on the MLLMs used to decompose complex instructions into simpler sub-instructions and corresponding regions, covering both open-source and proprietary models. As shown in Tab.~\ref{table4}, proprietary models demonstrate superior decomposition performance compared to open-source alternatives.

\begin{table}[ht]
\centering

\resizebox{\linewidth}{!}{  
\input{table/Ablation_MLLM}
}
\caption{Quantitative ablation results of different MLLMs.}
\label{table4}
\end{table}


%% file: table/Ablation_module.tex
\begin{tabular}{cccccccc}
\toprule
 \textbf{SACA} & \textbf{BCCA} & \textbf{CLIP-I$\uparrow$} & \textbf{DINO-I$\uparrow$} & \textbf{L1$\downarrow$} & \textbf{L2$\downarrow$} & \textbf{IC$\uparrow$} & \textbf{BC$\uparrow$}\\
\midrule
 --            & --            & 0.8784 & 0.9691 & 0.0710 & 0.1032 & 0.4922 & 0.6270\\
\checkmark    & --   & 0.8566 & 0.9690 & 0.0886 & 0.1320 & 0.7932  & 0.6851  \\
\checkmark    & \checkmark    & \textbf{0.8899} & \textbf{0.9833}  & \textbf{0.0489} & \textbf{0.1029} & \textbf{0.8046} & \textbf{0.7831} \\
\bottomrule
\end{tabular}

%% file: table/Ablation_MLLM.tex
\begin{tabular}{ccccccc}
\toprule
\textbf{Method} & \textbf{CLIP-I$\uparrow$} & \textbf{DINO-I$\uparrow$} & \textbf{L1$\downarrow$} & \textbf{L2$\downarrow$} & \textbf{IC$\uparrow$} & \textbf{BC$\uparrow$}\\
\midrule
Qwen2.5VL 3B & 0.8763 & 0.9673 & 0.0715 & 0.1270 & 0.7403 & 0.7139 \\
Qwen2.5VL 7B & 0.8789 & 0.9753 & 0.0643 & 0.1111 & 0.7505  & 0.7377 \\
Qwen2.5VL 32B & 0.8801 & 0.9768 & 0.0587 & 0.1080 & 0.7773 & 0.7539 \\
GPT4-o & \textbf{0.8812} & \textbf{0.9786}  & \textbf{0.0565} & \textbf{0.1050} & \textbf{0.7873} & \textbf{0.7705} \\
\bottomrule
\end{tabular}

%% file: section/con.tex
\section{Conclusion}
In this paper, we present MCIE-E1, a novel model for complex instruction-based image editing that leverages MLLMs to decompose intricate instructions. To enhance instruction adherence and preserve background consistency, we introduce two key modules: spatial-aware cross-attention and background-consistent cross-attention. In addition, we construct MCIE, the first image editing dataset centered on complex instructions, featuring precise spatial annotations and rigorous human filtering. Finally, we propose CIE-Bench and two new evaluation metrics for assessing complex instruction-based image editing. Experimental results on CIE-Bench demonstrate that MCIE-E1 consistently outperforms prior state-of-the-art methods across both quantitative and qualitative evaluations.
